\documentclass{article}
\usepackage{spconf,amsmath,graphicx}

\usepackage{graphicx}
\usepackage{subcaption}
\usepackage{wrapfig}

\usepackage{booktabs}
\usepackage{pifont}
\usepackage{xcolor}

\usepackage{amsmath, amssymb}
\usepackage{graphicx}
\usepackage{algorithm}
\usepackage{algorithmic}
\usepackage{graphicx}
\usepackage{booktabs}       
\usepackage{amsfonts}       
\usepackage{nicefrac}       
\usepackage{microtype}      
\usepackage{xcolor}         
\usepackage{multirow}
\usepackage{amsmath}
\usepackage{caption}
\usepackage{makecell}
\usepackage{caption}

\newcommand{\cmark}{\textcolor{cyan}{\ding{51}}}  
\newcommand{\xmark}{\textcolor{orange}{\ding{55}}} 

\title{\textit{DeCorStory}: Gram-Schmidt Prompt Embedding Decorrelation for Consistent Storytelling}
\name{Ayushman Sarkar$^{1}$, Zhenyu Yu$^{2,*}$, Mohd Yamani Idna Idris$^{2}$}
\address{
${^1}$ Birbhum Institute of Engineering and Technology, 
${^2}$ Universiti Malaya\\
\tt \small ayushmansarkar123@gmail.com; yuzhenyuyxl@foxmail.com; yamani@um.edu.my
}

\begin{document}
%
\maketitle

\begin{abstract}
Maintaining visual and semantic consistency across frames is a key challenge in text-to-image storytelling. Existing training-free methods, such as One-Prompt-One-Story, concatenate all prompts into a single sequence, which often induces strong embedding correlation and leads to color leakage, background blending, and identity drift. We propose \textit{DeCorStory}, a training-free inference-time framework that explicitly reduces inter-frame semantic interference. \textit{DeCorStory} applies Gram--Schmidt prompt embedding decorrelation to orthogonalize frame-level semantics, followed by singular-value reweighting to strengthen prompt-specific information and identity-preserving cross-attention to stabilize character identity during diffusion. The method requires no model modification or fine-tuning and can be seamlessly integrated into existing diffusion pipelines. 
Experiments demonstrate consistent improvements in prompt--image alignment, identity consistency, and visual diversity, achieving state-of-the-art performance among training-free baselines. Code is available at: https://github.com/YuZhenyuLindy/DeCorStory
\end{abstract}


\section{Introduction}

In narrative generation tasks such as animation, visual storytelling, and video synthesis, maintaining consistency across multiple frames remains a fundamental challenge for text-to-image (T2I) diffusion models~\cite{tewel2024consistory}. A coherent story requires the model to preserve the identity, appearance, and style of the main subject across different scenes, while still adapting flexibly to changing backgrounds and actions described in text. Recent advances in diffusion-based generation~\cite{rombach2022high} and identity-preserving T2I methods~\cite{li2024photomaker} have greatly improved visual quality and controllability, making text-driven storytelling increasingly feasible. Training-free methods such as One-Prompt-One-Story (1Prompt1Story)~\cite{liu2025onepromptonestory} further reduce computational overhead by concatenating all frame prompts into a single input sequence, thereby leveraging the implicit contextual consistency of large language--vision models.

However, this simple concatenation strategy introduces a new source of degradation. Because multiple frames often describe the same subject under varying contexts, their token embeddings in the language space tend to be highly correlated, leading to semantic interference between frames~\cite{cherti2023reproducible}. Visual attributes from one scene may leak into another, causing blended backgrounds or drifting details. As the number of frames increases or as scene diversity grows, this correlation problem becomes more severe, limiting the scalability of existing training-free methods in complex or long-form narratives~\cite{tewel2024consistory}. Since prior approaches rely solely on the model’s implicit context understanding, they lack an explicit mechanism to disentangle frame semantics, resulting in inconsistent generation when scenes differ substantially.

To address this limitation, we propose \textit{DeCorStory}, a training-free framework for consistent multi-frame generation. After prompt concatenation, \textit{DeCorStory} applies Gram--Schmidt Prompt Embedding Decorrelation to explicitly orthogonalize frame-level embeddings, thereby reducing inter-frame semantic overlap while maintaining a shared identity subspace. On top of this, we integrate Singular-Value Reweighting (SVR) and Identity-Preserving Cross-Attention (IPCA)~\cite{liu2025onepromptonestory} to strengthen frame-specific semantics and reinforce subject consistency, respectively. The entire process operates purely at inference time without any additional training or fine-tuning, and can be seamlessly integrated into existing diffusion pipelines that rely on strong foundational vision encoders such as DINOv2~\cite{oquab2023dinov2}. 
Our main \textbf{contributions} are:
\begin{itemize}
    \item \textbf{Frame-Embedding Correlation Analysis:} We identify and analyze the correlation of frame-level embeddings in concatenated prompts, revealing it as a key factor underlying semantic interference.
    \item \textbf{DeCorStory Framework:} We propose a lightweight Gram--Schmidt-based decorrelation module that can be directly incorporated into diffusion models without modifying or retraining them.
    \item \textbf{Unified Inference Pipeline:} We combine embedding decorrelation, SVR, and IPCA into a cohesive training-free pipeline that achieves identity-preserving and frame-specific multi-frame generation.
\end{itemize}

\section{Related Work}

\subsection{Consistent text-to-image generation}
Subject and identity consistency has been extensively studied in personalized text-to-image generation. Training-based methods, such as Textual Inversion~\cite{gal2023textualinversion} and DreamBooth~\cite{ruiz2023dreambooth}, learn subject-specific embeddings or fine-tune diffusion models to preserve identity across generations. Subsequent works introduce architectural extensions or adapters, including BLIP-Diffusion, IP-Adapter, and related designs~\cite{ye2023ipadapter}, to better inject identity cues into diffusion models. While these approaches achieve strong consistency, they typically require additional training or model modification, which limits flexibility compared with inference-time methods.

\subsection{Training-free consistent storytelling}
More recent works explore training-free consistency for multi-frame storytelling. ConsiStory and related approaches propagate identity information through attention manipulation during denoising~\cite{tewel2024consistory,zhou2024storydiffusion}. 1Prompt1Story~\cite{liu2025onepromptonestory} instead concatenates all frame prompts into a single sequence and attributes consistency to contextual coherence, further enhanced by singular-value reweighting and identity-preserving cross-attention. Our work follows this training-free paradigm but addresses an overlooked issue: prompt concatenation induces strong correlations among frame-level embeddings, leading to semantic leakage across frames.

\subsection{Prompt and embedding manipulation}
A large body of work studies how text embeddings or attention can be manipulated to steer diffusion models without retraining, including prompt reweighting, Prompt-to-Prompt editing, and plug-and-play control~\cite{mokady2023plug}. Other studies analyze embedding structure or selectively suppress unwanted concepts~\cite{heng2024selective}. \textit{DeCorStory} falls into this line of research by explicitly decorrelating frame-level prompt embeddings via Gram--Schmidt orthogonalization and selectively amplifying frame-specific semantics, providing a lightweight embedding-space solution for consistent multi-frame storytelling.

\section{Method}

\begin{figure*}[t]
    \centering
    \includegraphics[width=1\linewidth]{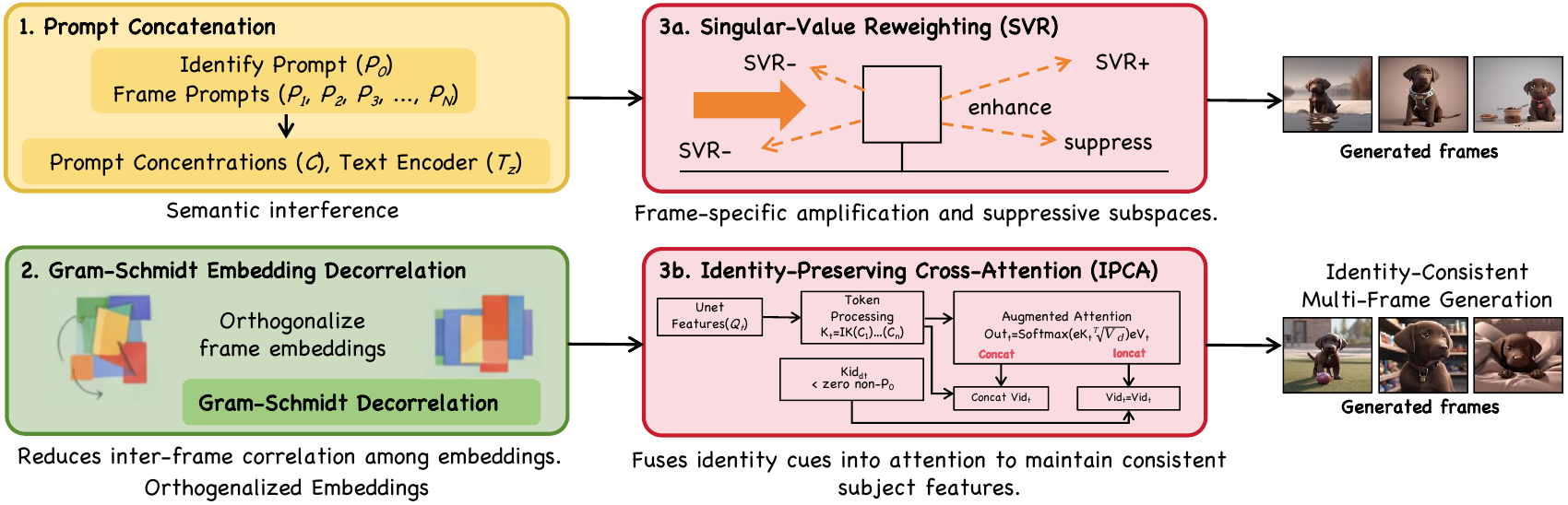}
    \caption{Overview of the \textit{DeCorStory} pipeline. The method consists of prompt concatenation, Gram–Schmidt embedding decorrelation, and inference-time modules (SVR and IPCA) for identity-consistent multi-frame generation. The identity prompt is shared across all frames, while decorrelated and reweighted frame-level embeddings provide clean, frame-specific conditioning.}
    \label{fig:overview}
\end{figure*}

\textit{DeCorStory} is a training-free inference-time framework that reduces inter-frame semantic interference via prompt concatenation, Gram--Schmidt decorrelation, and identity-aware reweighting on a frozen diffusion backbone (Figure~\ref{fig:overview}), requiring no model fine-tuning.

\begin{figure*}
\centering
\includegraphics[width=\linewidth]{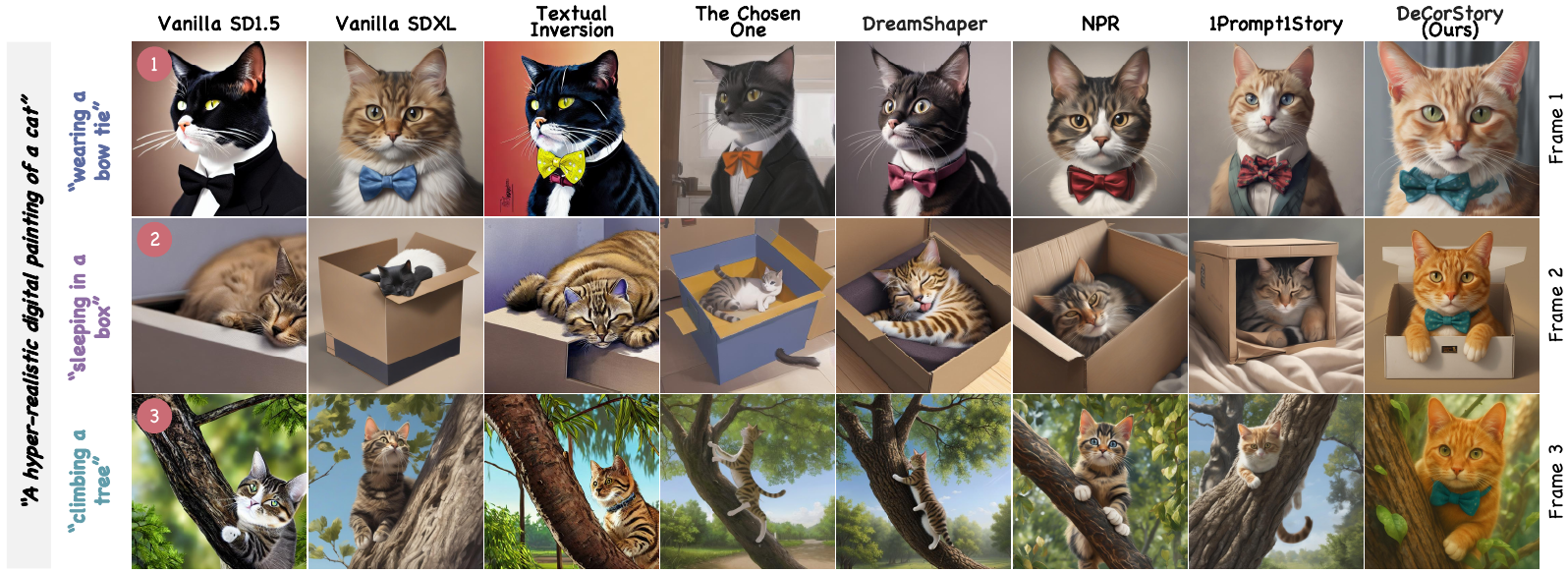}
    \caption{{Qualitative Comparison.} Visual comparison of different T2I generation methods on representative prompts. \textit{DeCorStory} produces more consistent global–local details, clearer character identity preservation, and better narrative coherence across scenes. More examples see Appendix Figure \ref{fig:supp1} and \ref{fig:supp2}.} 
    \label{fig_comparision}
\end{figure*}


\subsection{Prompt Concatenation}

Following 1Prompt1Story, we first form a single long prompt that concatenates the identity description and all frame-specific prompts.
This allows the model to share a consistent context across frames while retaining distinct scene information.
Concretely, the identity prompt summarizes the global subject (e.g., appearance, style), whereas each frame prompt describes local attributes such as pose, background, or events that are unique to that frame.
We concatenate the identity prompt $P_0$ with $N$ frame prompts $\{P_1,\dots,P_N\}$ and encode them as
\begin{equation}
C = \tau_{\xi}([P_0;P_1;\dots;P_N]),
\end{equation}
\begin{equation}
C = [c_{\mathrm{SOT}},\, c_{P_0},\, c_{P_1},\, \dots,\, c_{P_N},\, c_{\mathrm{EOT}}]
\in \mathbb{R}^{M\times D},
\end{equation}
where $\tau_{\xi}(\cdot)$ denotes the text encoder, $M$ is the total number of tokens, and $D$ is the embedding dimension.
Here $c_{\mathrm{SOT}}$ and $c_{\mathrm{EOT}}$ are the special start-of-text and end-of-text tokens, and each $c_{P_k}$ corresponds to the embedding of the (possibly multi-token) description for frame $k$ within the joint prompt context.
We collect the frame-level token embeddings as
\begin{equation}
X = 
\begin{bmatrix}
c_{P_1}\\
c_{P_2}\\
\vdots\\
c_{P_N}
\end{bmatrix}
\in \mathbb{R}^{N\times D}, \quad
c_{P_k}\in\mathbb{R}^{1\times D}.
\end{equation}
This matrix $X$ can be interpreted as a compact representation of all frame prompts in a shared semantic space.
However, because all $c_{P_k}$ are encoded jointly with the same identity prompt $P_0$, they tend to be highly correlated, which can cause different frames to “bleed” into each other during generation.
The next stage explicitly addresses this issue.

\subsection{Gram--Schmidt Embedding Decorrelation}

To mitigate inter-frame correlation, we perform row-wise Gram--Schmidt orthogonalization on $X$ to obtain $\tilde{X}$:
\begin{align}
\tilde{c}_{P_1} &= \frac{c_{P_1}}{\|c_{P_1}\|_2},\\
\tilde{c}_{P_k} &= 
\frac{c_{P_k} - \sum_{i=1}^{k-1} \langle c_{P_k},\, \tilde{c}_{P_i}\rangle\, \tilde{c}_{P_i}}
{\left\|c_{P_k} - \sum_{i=1}^{k-1} \langle c_{P_k},\, \tilde{c}_{P_i}\rangle\, \tilde{c}_{P_i}\right\|_2},
\quad k=2,\dots,N,
\end{align}
where inner product $\langle a,b\rangle = a\,b^\top$ for row vectors.
Geometrically, we treat each $c_{P_k}$ as a vector in the $D$-dimensional embedding space and construct an orthonormal basis that spans the same subspace but with pairwise-orthogonal frame directions.
We then construct the decorrelated token matrix
\begin{equation}
\tilde{C} = [\,c_{\mathrm{SOT}},\, c_{P_0},\, \tilde{c}_{P_1},\, \dots,\, \tilde{c}_{P_N},\, c_{\mathrm{EOT}}\,].
\end{equation}
Compared to $C$, the identity embedding $c_{P_0}$ and special tokens remain unchanged, while only the frame-level embeddings are rotated into orthogonal directions.
This preserves the overall semantic content of each prompt but reduces overlap between different frames in the shared embedding space.


We use a numerically stable modified Gram--Schmidt procedure, whose ordering has negligible impact in practice, and whose computational cost is marginal given the small number of frames. This operation encourages approximate orthogonality among frame-level embeddings, reducing inter-frame semantic entanglement before diffusion inference.

\subsection{Integration with SVR and IPCA}


For each target frame $j$, SVR enhances $P_j$ while attenuating other frames, followed by identity reinforcement via IPCA. This conditioning is applied on-the-fly during sampling with a frame-specific text matrix, requiring no architectural changes or additional training.

\subsubsection{Singular-Value Reweighting (SVR).}

Let the \emph{express} set be $X^{\mathrm{exp}}=[\tilde{c}_{P_j};\, c_{\mathrm{EOT}}]\in\mathbb{R}^{2\times D}$.
Compute SVD on its column space (equivalently on $(X^{\mathrm{exp}})^\top$):
\begin{equation}
(X^{\mathrm{exp}})^\top = U\,\Sigma\,V^\top,
\end{equation}
\begin{equation}
\Sigma=\mathrm{diag}(\sigma_1,\dots,\sigma_r),\ \sigma_1\ge\dots\ge\sigma_r\ge 0.
\end{equation}
Here the singular vectors capture principal semantic directions jointly supported by the current frame and the end-of-text embedding.
Reweighting the singular values therefore allows us to emphasize or suppress entire semantic subspaces rather than individual coordinates.
SVR$+$ amplifies singular values:
\begin{equation}
\hat{\sigma}_\ell = \beta\, e^{\alpha\,\sigma_\ell}\,\sigma_\ell,\quad \forall \ell,
\end{equation}
yielding $\hat{\Sigma}=\mathrm{diag}(\hat{\sigma}_\ell)$ and the updated tokens
\begin{equation}
\hat{X}^{\mathrm{exp}} = \big(U\,\hat{\Sigma}\,V^\top\big)^\top = [\,\hat{c}_{P_j};\, \hat{c}_{\mathrm{EOT}}\,].
\end{equation}
The hyperparameters control how aggressively dominant singular directions are boosted: larger values lead to stronger emphasis on the most salient semantics of frame $j$.
For each \emph{suppress} frame $k\neq j$, form $X^{\mathrm{sup}}_k=[\tilde{c}_{P_k};\, \hat{c}_{\mathrm{EOT}}]$ and compute SVD similarly to get $\hat{\Sigma}_k$.
SVR$-$ attenuates singular values:
\begin{equation}
\tilde{\sigma}_{k,\ell} = \beta'\, e^{-\alpha'\,\hat{\sigma}_{k,\ell}}\ \hat{\sigma}_{k,\ell},\quad \forall \ell,
\end{equation}
leading to $\tilde{X}^{\mathrm{sup}}_k=[\,\tilde{c}_{P_k}^{\,\downarrow};\, \tilde{c}_{\mathrm{EOT}}\,]$. 
Assemble the per-frame conditioned text matrix:
\begin{equation}
\tilde{C}^{(j)} = [\,c_{\mathrm{SOT}},\, c_{P_0},\, \tilde{c}_{P_1}^{\,\downarrow},\dots,\hat{c}_{P_j},\dots,\tilde{c}_{P_N}^{\,\downarrow},\, \tilde{c}_{\mathrm{EOT}}\,].
\end{equation}
Intuitively, SVR$+$ emphasizes the semantic subspace of the current frame, while SVR$-$ suppresses the influence of other frames, yielding cleaner frame-specific conditioning.
Because SVR operates on a low-rank factorization of the token matrix, it provides a compact and interpretable way to modulate frame semantics without directly editing the raw embeddings.

\subsubsection{Identity-Preserving Cross-Attention (IPCA).}

At denoising step $t$, the cross-attention uses queries $Q_t$ from the UNet features and keys/values from $\tilde{C}^{(j)}$:
\begin{equation}
K_t = \ell_K(\tilde{C}^{(j)}),\quad V_t=\ell_V(\tilde{C}^{(j)}).
\end{equation}
Let $K_t^{\mathrm{id}}, V_t^{\mathrm{id}}$ be copies where all non-identity (non-$P_0$) token features are zeroed.
We concatenate
\begin{equation}
\widetilde{K}_t = \mathrm{Concat}(K_t,\ K_t^{\mathrm{id}}),\qquad 
\widetilde{V}_t = \mathrm{Concat}(V_t,\ V_t^{\mathrm{id}}),
\end{equation}
and compute the augmented attention
\begin{equation}
\widetilde{A}_t=\mathrm{softmax}\Big(\frac{Q_t\,\widetilde{K}_t^\top}{\sqrt{d}}\Big),\qquad 
\mathrm{Out}_t=\widetilde{A}_t\,\widetilde{V}_t,
\end{equation}
where $d$ is the key/query dimension.
The original keys and values encode mixed semantics, while an identity-only branch reserves attention capacity for foreground identity cues.


The attention output $\mathrm{Out}_t$ combines identity cues and frame-specific semantics, enabling stable cross-frame consistency when applied at each denoising step alongside decorrelated and SVR-modulated prompts.

\begin{table*}
\centering
\caption{Quantitative comparison. The best and second best results are highlighted in \textbf{bold} and \underline{underline}, respectively. Vanilla SD1.5 and Vanilla SDXL are shown as references and excluded from this comparison.}
\resizebox{\linewidth}{!}{
\begin{tabular}{lcccccccc}
\toprule
\textbf{Method} & \textbf{Base Model} & \textbf{Train-Free} & \textbf{CLIP-T$\uparrow$} & \textbf{CLIP-I$\uparrow$} & \textbf{DreamSim$\downarrow$} & \textbf{Steps} & \textbf{Memory (GB)$\downarrow$} & \textbf{Inference Time (s)$\downarrow$} \\
\midrule
Vanilla SD1.5 & - & - & 0.8353 & 0.7474 & 0.5873 & 50 & 4.73 & 2.4657 \\
Vanilla SDXL & - & - & 0.9074 & 0.8165 & 0.5292 & 50 & 16.04 & 13.0890 \\
\midrule
BLIP-Diffusion & SD1.5 & \xmark & 0.7607 & 0.8863 & 0.2830 & 26 & \textbf{7.75} & \textbf{1.9284} \\
Textual Inversion & SD1.5 & \xmark & 0.8378 & 0.8229 & 0.4268 & 40 & 32.94 & 282.507 \\
The Chosen One & SDXL & \xmark & 0.7614 & 0.7831 & 0.4929 & 35 & \underline{10.93} & \underline{11.2073} \\
PhotoMaker & SDXL & \xmark & 0.8651 & 0.8465 & 0.3996 & 50 & 23.79 & 18.0259 \\
IP-Adapter & SDXL & \xmark & 0.8458 & \textbf{0.9429} & \textbf{0.1462} & 30 & 19.39 & 13.4594 \\
\midrule
ConsiStory & SDXL & \cmark & 0.8769 & 0.8737 & 0.3188 & 50 & 34.55 & 34.5894 \\
StoryDiffusion & SDXL & \cmark & 0.8877 & 0.8755 & 0.3212 & 50 & 45.61 & 25.6928 \\
Naive Prompt Reweighting (NPR) & SDXL & \cmark & 0.8411 & 0.8916 & 0.2548 & 50 & 16.04 & 17.2413 \\
1Prompt1Story & SDXL & \cmark & \underline{0.8942} & 0.9117 & 0.1993 & 50 & 18.70 & 23.2088 \\
\textbf{\textit{DeCorStory (Ours)}} & SDXL & \cmark & \textbf{0.9001} & \underline{0.9134} & \underline{0.1922} & 50 & 18.72 & 23.3521 \\
\bottomrule
\end{tabular}
}
\label{tab_quantitative_comparison}
\end{table*}

\begin{table*}
\centering
\caption{User study with 100 participants voting for the best consistent T2I generation method according to human preference.}
\resizebox{1.0\linewidth}{!}{
\begin{tabular}{ccccccccc}
\toprule
\textbf{Method} & \textbf{Vanilla SD1.5} & \textbf{Vanilla SDXL} & \textbf{Textual Inversion} & \textbf{The Chosen One} & \textbf{DreamShaper} & \textbf{NPR}
 & \textbf{1Prompt1Story} & \textbf{\textit{DeCorStory}} \\
\midrule
 (\%)$\uparrow$ & 2 & 4 & 8 & 10 & 16 & 18 & \underline{19} & \textbf{23} \\
\bottomrule
\end{tabular}
}
\label{tab_user_study}
\end{table*}

\begin{table}
\centering
\caption{Ablation study. We evaluate the contribution of each component, including the GS, SVR$^{+}$/SVR$^{-}$, and IPCA.}
\resizebox{0.95\linewidth}{!}{
\begin{tabular}{lccc}
\toprule
\textbf{Method} & \textbf{CLIP-T$\uparrow$} & \textbf{CLIP-I$\uparrow$} & \textbf{DreamSim$\downarrow$} \\
\midrule
w/o GS          & 0.8874 & 0.9021 & 0.2375 \\
w/o SVR         & 0.8896 & 0.9058 & 0.2312 \\
w/o IPCA        & 0.8912 & 0.9084 & 0.2259 \\
Only GS         & 0.8825 & 0.8993 & 0.2418 \\
Only SVR        & 0.8859 & 0.9046 & 0.2356 \\
GS + SVR        & \underline{0.8926} & 0.9091 & \underline{0.2237} \\
GS + IPCA       & 0.8918 & \underline{0.9102} & 0.2241 \\
\textbf{Full (GS + SVR + IPCA)} & \textbf{0.9001} & \textbf{0.9134} & \textbf{0.1922} \\
\bottomrule
\end{tabular}
}
\label{tab_ablation}
\end{table}

\section{Experiment}

\subsection{Implementation Details}
\noindent\textbf{Dataset Description.} 
In our experiments, we adopt the ConsiStory+ dataset \cite{liu2025onepromptonestory}, which is an extended version of the original ConsiStory benchmark \cite{tewel2024training}.
It contains a broader range of subjects, scene descriptions, and visual styles, making it suitable for evaluating multi-frame and narrative generation tasks.
The dataset includes 200 prompt sets with approximately 1500 generated frames, on which we evaluate both prompt alignment and identity consistency.

\noindent\textbf{Experimental Settings.} 
We compare \textit{DeCorStory} with several representative text-to-image consistency generation methods, including 1Prompt1Story \cite{liu2025onepromptonestory}, BLIP-Diffusion \cite{li2023blip}, Textual Inversion (TI) \cite{galimage}, IP-Adapter \cite{ye2023ip}, PhotoMaker \cite{li2024photomaker}, The Chosen One \cite{avrahami2024chosen}, ConsiStory \cite{tewel2024training}, and StoryDiffusion \cite{zhou2024storydiffusion}.
All baselines are evaluated using their official configurations or publicly released implementations to ensure fairness.

\noindent\textbf{Evaluation Metrics.} 
We evaluate our method using three complementary metrics.
(1) Prompt Alignment. We compute the average CLIP Score \cite{hessel2021clipscore} between each generated frame and its corresponding prompt, denoted as CLIP-T, to measure the semantic alignment between text and image.
(2) Identity Consistency. We adopt both DreamSim \cite{fu2023dreamsim} and CLIP-I \cite{hessel2021clipscore} to quantify visual consistency across frames. DreamSim reflects perceptual similarity, while CLIP-I is computed as the cosine similarity between CLIP image embeddings.
(3) Background Neutralization. Following Fu et al. \cite{fu2023dreamsim}, we remove backgrounds using CarveKit \cite{selin2023carvekit} and replace them with random noise to ensure that the similarity evaluation focuses solely on subject identity rather than contextual artifacts.

\subsection{Comparison}



\noindent\textbf{Qualitative Comparison.}
Figure~\ref{fig_comparision} shows qualitative results on two 3-frame stories with different styles and actions. Vanilla SD1.5/SDXL and existing personalization or storytelling baselines often exhibit identity drift, style inconsistency, and missing semantic elements across frames, such as changes in appearance, accessories, or background details. In contrast, \emph{DeCorStory} generates a single recognizable character that remains consistent across all frames while accurately following the narrative actions. Visual attributes, accessories, and rendering style are preserved, and the background context evolves coherently with the story, demonstrating improved identity preservation, style coherence, and story-level faithfulness over competing methods.

\noindent\textbf{Quantitative Comparison.}
Table~\ref{tab_quantitative_comparison} compares \textit{DeCorStory} with both training-based and training-free baselines on ConsiStory$+$. Without any additional training or fine-tuning, \textit{DeCorStory} achieves comparable or superior performance to training-based methods and ranks first among training-free approaches, attaining the highest CLIP-T (0.9001) and CLIP-I (0.9134) while achieving the lowest DreamSim (0.1922). Compared with training-free storytelling baselines, including ConsiStory, StoryDiffusion, and 1Prompt1Story, \textit{DeCorStory} consistently improves all metrics, confirming that explicit embedding decorrelation effectively suppresses inter-frame semantic interference. Notably, the gains over 1Prompt1Story demonstrate measurable benefits beyond simple prompt concatenation, while incurring negligible additional computational overhead at inference time.

\subsection{User Study}
We conduct a user study to evaluate perceptual quality and consistency across eight methods, including vanilla diffusion baselines, personalization approaches, and training-free storytelling models. Following ConsiStory$+$, 30 prompt sets with four frames each are used to generate image sequences, and 100 participants select the sequence that best balances identity consistency, prompt alignment, and visual diversity. As reported in Table~\ref{tab_user_study}, \textit{DeCorStory} receives the highest preference rate, outperforming both personalization methods and training-free baselines such as NPR and 1Prompt1Story. This result is consistent with our quantitative evaluation and indicates that explicit embedding decorrelation combined with SVR and IPCA produces image sequences that are more coherent and semantically faithful from a human perspective. 

\subsection{Ablation Study}




Table~\ref{tab_ablation} reports an ablation study analyzing the contributions of GS, SVR, and IPCA. Removing GS leads to clear drops in CLIP-T and DreamSim, indicating increased inter-frame semantic interference, while disabling SVR reduces identity consistency by weakening the balance between shared and frame-specific semantics. Eliminating IPCA further degrades cross-frame identity preservation, particularly under appearance variations. 
Combining GS and SVR already yields stable scene-level semantics, but the full configuration (GS + SVR + IPCA) achieves the best performance across all metrics (CLIP-T = 0.9001, CLIP-I = 0.9134, DreamSim = 0.1922). These results demonstrate that embedding decorrelation, semantic reweighting, and identity-aware attention are complementary, enabling \textit{DeCorStory} to achieve strong prompt alignment and identity consistency.

\section{Conclusion}



We presented \textit{DeCorStory}, a training-free framework that enhances multi-frame storytelling by reducing semantic interference among frame-level prompts. Through Gram--Schmidt Prompt Embedding Decorrelation (GS), Singular-Value Reweighting (SVR), and Identity-Preserving Cross-Attention (IPCA), our method disentangles correlated prompt embeddings, amplifies frame-specific semantics, and stabilizes identity during diffusion. Operating entirely at inference time without model retraining, \textit{DeCorStory} achieves superior prompt--image alignment, identity consistency, and narrative coherence compared with existing training-free baselines, highlighting embedding decorrelation as an effective principle for text-to-image storytelling.

\clearpage

\bibliographystyle{IEEEbib}
\bibliography{main}

\clearpage
\appendix
\onecolumn

\setcounter{page}{1}
\setcounter{section}{0}
\setcounter{figure}{0}
\renewcommand{\thetable}{A\arabic{table}}
\renewcommand{\thefigure}{A\arabic{figure}}

\section{Appendix}

\begin{algorithm}[t]
\caption{\textit{DeCorStory} Inference Pipeline}
\label{alg:decorstory}
\begin{algorithmic}[1]
\STATE \textbf{Input:} Prompts $\{P_0, P_1, \dots, P_N\}$; encoder $\tau_\xi$; UNet $\epsilon_\theta$; decoder $D$; diffusion steps $T$; SVR parameters $(\alpha, \beta, \alpha', \beta')$
\STATE \textbf{Output:} Generated frames $\{I_j\}_{j=1}^N$

\vspace{2pt}
\STATE \textcolor{gray}{{Stage 1: Prompt Encoding and Concatenation}}
\STATE $C \leftarrow \tau_{\xi}([P_0; P_1; \dots; P_N])$
\STATE \hspace{1.2em} $= [c_{\mathrm{SOT}}, c_{P_0}, c_{P_1}, \dots, c_{P_N}, c_{\mathrm{EOT}}]$

\vspace{2pt}
\STATE \textcolor{gray}{{Stage 2: Gram--Schmidt Decorrelation}}
\STATE $\tilde{c}_{P_1} \leftarrow c_{P_1}/\|c_{P_1}\|_2$
\FOR{$k=2$ \TO $N$}
    \STATE $\tilde{c}_{P_k} \leftarrow 
    \dfrac{c_{P_k} - \sum_{i<k}\langle c_{P_k}, \tilde{c}_{P_i}\rangle \tilde{c}_{P_i}}
          {\bigl\|c_{P_k} - \sum_{i<k}\langle c_{P_k}, \tilde{c}_{P_i}\rangle \tilde{c}_{P_i}\bigr\|_2}$ 
\ENDFOR
\STATE $\tilde{C} \leftarrow [c_{\mathrm{SOT}}, c_{P_0}, \tilde{c}_{P_1}, \dots, \tilde{c}_{P_N}, c_{\mathrm{EOT}}]$

\vspace{2pt}
\FOR{$j=1$ \TO $N$}
    \STATE \textcolor{gray}{{Stage 3: SVR Reweighting for Frame $j$}}
    \STATE $X^{\mathrm{exp}} \leftarrow [\tilde{c}_{P_j}; c_{\mathrm{EOT}}]$;\quad $(X^{\mathrm{exp}})^\top = U\Sigma V^\top$
    \STATE $\hat{\Sigma} \leftarrow \mathrm{diag}(\hat{\sigma}_\ell)$ where $\hat{\sigma}_\ell = \beta e^{\alpha\sigma_\ell}\sigma_\ell$
    \STATE $[\hat{c}_{P_j}; \hat{c}_{\mathrm{EOT}}] \leftarrow (U\hat{\Sigma}V^\top)^\top$
    \FOR{each $k \neq j$}
        \STATE $X^{\mathrm{sup}}_k \leftarrow [\tilde{c}_{P_k}; \hat{c}_{\mathrm{EOT}}]$;\quad $(X^{\mathrm{sup}}_k)^\top = U_k\hat{\Sigma}_kV_k^\top$
        \STATE $\tilde{\Sigma}_k \leftarrow \mathrm{diag}(\tilde{\sigma}_{k,\ell})$ where $\tilde{\sigma}_{k,\ell} = \beta' e^{-\alpha'\hat{\sigma}_{k,\ell}}\hat{\sigma}_{k,\ell}$
        \STATE $[\tilde{c}_{P_k}^{\downarrow}; \tilde{c}_{\mathrm{EOT}}] \leftarrow (U_k\tilde{\Sigma}_kV_k^\top)^\top$
    \ENDFOR
    \STATE $\tilde{C}^{(j)} \leftarrow [c_{\mathrm{SOT}}, c_{P_0}, \tilde{c}_{P_1}^{\downarrow}, \dots, \hat{c}_{P_j}, \dots, \tilde{c}_{P_N}^{\downarrow}, \tilde{c}_{\mathrm{EOT}}]$

\vspace{2pt}
    \STATE \textcolor{gray}{{Stage 4: Denoising with IPCA}}
    \STATE Initialize $z_T \sim \mathcal{N}(0, I)$
    \FOR{$t=T$ \TO $1$}
        \STATE $K_t \leftarrow \ell_K(\tilde{C}^{(j)})$;\quad $V_t \leftarrow \ell_V(\tilde{C}^{(j)})$
        \STATE $K_t^{\mathrm{id}} \leftarrow K_t$;\quad $V_t^{\mathrm{id}} \leftarrow V_t$
        \STATE Zero all non-$P_0$ token rows in $K_t^{\mathrm{id}}, V_t^{\mathrm{id}}$
        \STATE $\widetilde{K}_t \leftarrow \mathrm{Concat}(K_t, K_t^{\mathrm{id}})$;\quad $\widetilde{V}_t \leftarrow \mathrm{Concat}(V_t, V_t^{\mathrm{id}})$
        \STATE $\mathrm{Out}_t \leftarrow \mathrm{softmax}(Q_t\widetilde{K}_t^\top/\sqrt{d}) \,\widetilde{V}_t$
        \STATE $z_{t-1} \leftarrow \epsilon_\theta(z_t, t, \tilde{C}^{(j)}; \mathrm{Out}_t)$
    \ENDFOR
    \STATE $I_j \leftarrow D(z_0)$
\ENDFOR
\STATE \textbf{return} $\{I_j\}_{j=1}^N$
\end{algorithmic}
\end{algorithm}

\begin{figure*}[h]
\centering
\includegraphics[width=\linewidth]{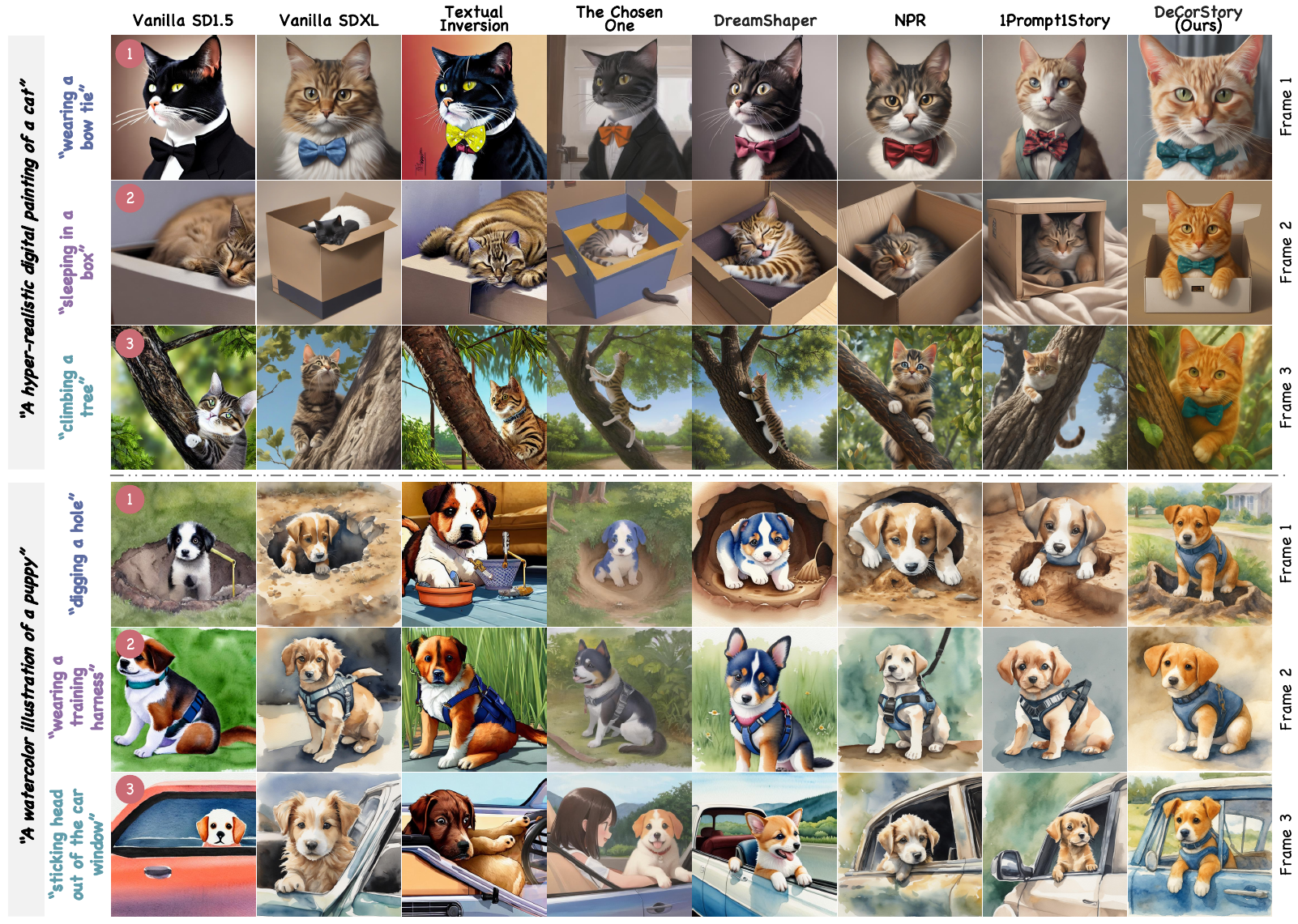}
    \caption{Qualitative comparison results.}
    \label{fig:supp1}
\end{figure*}

\begin{figure*}[h]
    \centering
    \includegraphics[width=1\linewidth]{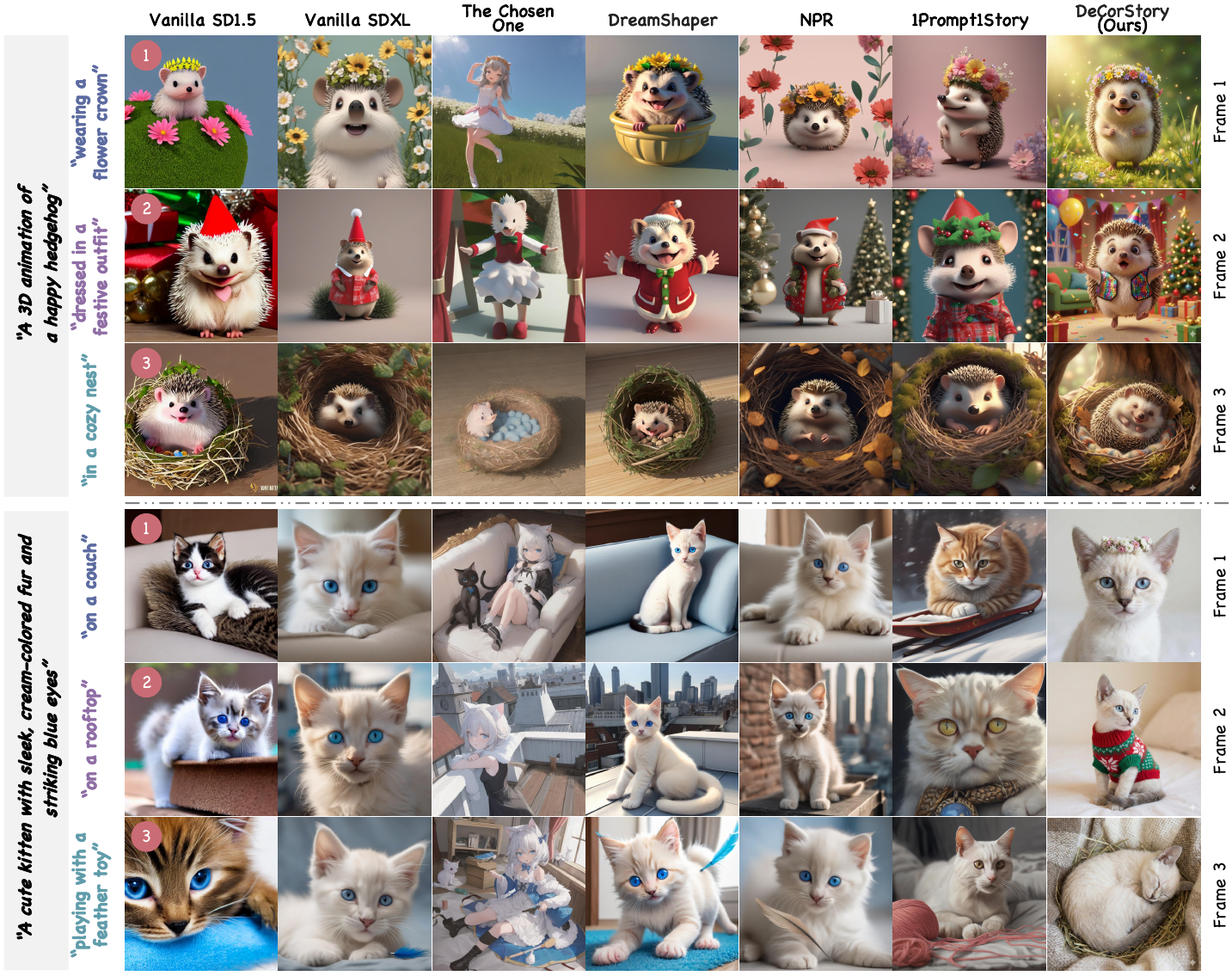}
    \caption{Qualitative comparison results.}
    \label{fig:supp2}
\end{figure*}

\end{document}